\documentclass[letterpaper]{article}
\usepackage{aaai25}  
\usepackage{times}
\usepackage{helvet}
\usepackage{courier}
\usepackage[hyphens]{url}
\usepackage{graphicx}
\usepackage{amsmath,amssymb}
\usepackage{booktabs}
\usepackage{natbib}
\usepackage{caption}
\usepackage{amsmath, amssymb, amsfonts, mathtools}
\usepackage{booktabs}
\usepackage{algorithm}
\usepackage{algpseudocode}
\usepackage{xcolor}
\usepackage{enumitem}
\usepackage{multirow}
\usepackage{tikz}
\usepackage{longtable}
\usepackage{comment}
\usepackage{pifont}  
\usepackage{graphicx}
\usepackage{subcaption}
\usepackage{xcolor}
\definecolor{RedOrange}{RGB}{255, 83, 73}  

\newcounter{ToDo}
\newcounter{gaocomm} 
\newcounter{Note}
\definecolor{blue-violet}{rgb}{0.00,0.75,0.90}
\definecolor{mygreen}{rgb}{0.0, 0.5, 0.0}
\definecolor{awesome}{rgb}{1.0, 0.13, 0.32}
\definecolor{bostonuniversityred}{rgb}{1.0, 0.0, 0.0}

\newcommand{\cmark}{\ding{51}} 
\newcommand{\xmark}{\ding{55}} 
\newcommand{\semitick}{{\cmark}\kern-0.62em{\xmark}}

\usepackage{amsthm} 

\theoremstyle{definition}
\newtheorem{definition}{Definition}

\title{STPFormer: A State-of-the-Art Pattern-Aware Spatio-Temporal Transformer for Traffic Forecasting}
\author{
    Jiayu Fang, Jessica Shao,  Boris Choy and Junbin Gao
}
\affiliations {
    University of Somewhere \\
    \texttt{jiayu.fang@sydney.edu.au}
}

\begin{document}
\maketitle




\begin{abstract}
Spatio-temporal traffic forecasting is challenging due to complex temporal patterns, dynamic spatial structures, and diverse input formats. Although Transformer-based models offer strong global modeling, they often struggle with rigid temporal encoding and weak space-time fusion. We propose STPFormer, a Spatio-Temporal Pattern-Aware Transformer that achieves state-of-the-art performance via unified and interpretable representation learning. It integrates four modules: Temporal Position Aggregator (TPA) for pattern-aware temporal encoding, Spatial Sequence Aggregator (SSA) for sequential spatial learning, Spatial-Temporal Graph Matching (STGM) for cross-domain alignment, and an Attention Mixer for multi-scale fusion. Experiments on five real-world datasets show that STPFormer consistently sets new SOTA results, with ablation and visualizations confirming its effectiveness and generalizability.

\end{abstract}

\section{Introduction}
\label{introduction}

With the increasing complexity of transportation systems and diversified road networks, addressing congestion and optimizing routes have become urgent priorities. Although various models have been proposed to tackle these challenges, there is still no unified solution with sufficient generalizability. As a result, prediction errors persist, leading to inefficient fuel consumption, labor misallocation, and suboptimal infrastructure investment, ultimately causing significant economic losses.

 Classical statistical approaches like the Vector AutoRegressive model \cite{Luetkepohl2005} and Support Vector Regression \cite{Smola2004} are interpretable but often rely on restrictive assumptions (e.g., stationarity, node independence) and struggle with nonlinear and spatial dependencies \cite{ZHANG199835,Wu2004SVR}. In recent years, Graph Neural Networks (GNNs) have gained attention for modeling spatial relations through message passing (e.g., STGCN, DCRNN, GWNET) \cite{ijcai2018p505,LiYS018,wu2019graph}. Yet, most still depend on fixed adjacency matrices and local aggregation, which can limit their ability to capture long-range or evolving connections.

To overcome these shortcomings, researchers have proposed attention-based architectures. Transformer variants such as STTN \cite{s24175502}, GMAN \cite{ZhengFW020}, MTGNN \cite{wu2020}, and PDFormer \cite{10.1609/aaai.v37i4.25556} use self-attention to learn global dependencies across time and space. However, many still face challenges with rigid time encoding, static spatial assumptions, or weak spatiotemporal integration. Other lines of work, including CNN-based methods (e.g., ST-ResNet) \cite{Zhang2016}, ODE-inspired designs (e.g., STGODE, STGNCDE) \cite{3467430,3604808}, and hybrid models, have made progress but often lack consistency or a unified representation approach.

These models all perform well when carrying out traffic prediction tasks. However, they have a fatal problem, that is, they do not have a strong generalization ability and can quickly adapt to various traffic models in different cities with minimal adjustments. To address these challenges, our team proposed STPFormer, which is a spatio-temporal pattern-aware converter that simulates complex spatio-temporal dependencies through a unified and modular design. This architecture jointly considers temporal evolution, spatial structure and their cross-domain interaction.

To operationalize these patterns, STPFormer integrates four dedicated modules:
\begin{itemize}
    \item \textbf{Temporal Position Aggregator (TPA)}: introduces learnable temporal encodings via random-walk embeddings and pattern-key attention, enabling the model to capture diverse and long-range temporal structures.
    \item \textbf{Spatial Sequence Aggregator (SSA)}: formulates spatial modeling as a sequential task, using LSTM-augmented attention to learn dynamic, order-aware spatial interactions.
    \item \textbf{Spatial-Temporal Graph Memory (STGM)}: facilitates bidirectional alignment between spatial and temporal features, enhancing context propagation and temporal precision.
    \item \textbf{Attention Mixer}: aggregates hierarchical representations from all encoder layers using learnable projection and residual fusion, forming a unified feature space for downstream forecasting.
\end{itemize}

We validate STPFormer on five benchmark datasets, and the experimental results show that, compared with the competing baselines, STPFormer achieves the most advanced performance and can adapt to various types of data without adjusting parameters. Its accuracy far exceeds that of the existing sota models.

\section{Preliminaries}
\label{sec:preliminaries}
This section introduces the foundational concepts and mathematical formulations underlying our traffic flow prediction framework.

\subsection{Problem Formulation}
\label{subsec:problem_formulation}

\begin{definition}[Road Network as a Graph]
\label{def:graph_representation}
A road network is modeled as a directed graph $\mathcal{G} = (\mathcal{V}, \mathcal{E}, \mathbf{A})$, where $\mathcal{V} = \{v_1, \dots, v_N\}$ is the set of $N$ nodes (e.g., sensors or intersections), $\mathcal{E} \subseteq \mathcal{V} \times \mathcal{V}$ is the set of directed edges, and $\mathbf{A} \in \{0,1\}^{N \times N}$ is the adjacency matrix such that $\mathbf{A}_{ij} = 1$ if there is a directed edge from $v_i$ to $v_j$.
\end{definition}

\begin{definition}[Spatio-Temporal Traffic Flow Representation]
\label{def:spatiotemporal_representation}
For a road network $\mathcal{G}$ with $N$ nodes, the traffic flow at time step $t$ is represented as $\mathbf{X}_t \in \mathbb{R}^{N \times d}$, where $d$ denotes the number of traffic features (e.g., speed, volume, occupancy). Over a time horizon $T$, the complete traffic sequence is encoded as a third-order tensor:
\begin{equation}
\mathcal{X} =\mathbf{X}_{1:T} :=[\mathbf{X}_1, \mathbf{X}_2, \ldots, \mathbf{X}_T] \in \mathbb{R}^{T \times N \times d},
\end{equation}
where the dimensions correspond to time, space, and feature channels, respectively.
\end{definition}

\paragraph{Problem Statement}
Given historical observations $\mathbf{X}_{t-m+1:t} \in \mathbb{R}^{m \times N \times d}$ of $m$ look-back steps at time $t$ over the road network $\mathcal{G}$, the goal is to predict traffic states for the next $h$ steps, yielding $\hat{\mathbf{X}}_{t+1:t+h} \in \mathbb{R}^{h \times N \times d}$. Formally, the task is to learn a function $f$ that maps historical traffic and topology to future traffic:
\begin{equation}
[\hat{\mathbf{X}}_{t+1}, \dots, \hat{\mathbf{X}}_{t+h}]=f([\mathbf{X}_{t-m+1}, \dots, \mathbf{X}_t], \mathcal{G}),
\label{eq:prediction}
\end{equation}

\subsection{Attention Mechanism}
\label{subsec:attention_mechanism}
To support our attention-based architecture, we first review the canonical self-attention mechanism, known as scaled dot-product attention. The attention was originally developed for sequence modeling~\cite{VAS}. Given input vectors $\mathbf{X}\!\in\!\mathbb{R}^{N\times d}$, queries ($\mathbf{Q}$), keys ($\mathbf{K}$), and values ($\mathbf{V}$) can be formulated as:
\begin{equation}
\mathbf{Q}= \mathbf{X}\mathbf{W}_Q,\quad
\mathbf{K}= \mathbf{X}\mathbf{W}_K,\quad
\mathbf{V}= \mathbf{X}\mathbf{W}_V,
\end{equation}
where $\mathbf{W}_{Q,K,V}\!\in\!\mathbb{R}^{d\times d_*}$ are learnable parameters for queries, keys, and values. Scaled dot products yield scores and weights:
\begin{equation}
\mathbf{A}= \text{softmax}\!\Bigl(\tfrac{\mathbf{Q}\mathbf{K}^\top}{\sqrt{d_k}}\Bigr)\in\mathbb{R}^{N\times N},
\end{equation}
and the final output is
\begin{equation}
\mathbf{Y}= \mathbf{A}\mathbf{V}\in\mathbb{R}^{N\times d_v}.
\end{equation}

\section{Methodology}
\label{sec:methodology}
\subsection{Framework Overview}
\begin{figure}[htbp]
\mbox{}\hspace{-0.65cm}\includegraphics[scale = 0.19]{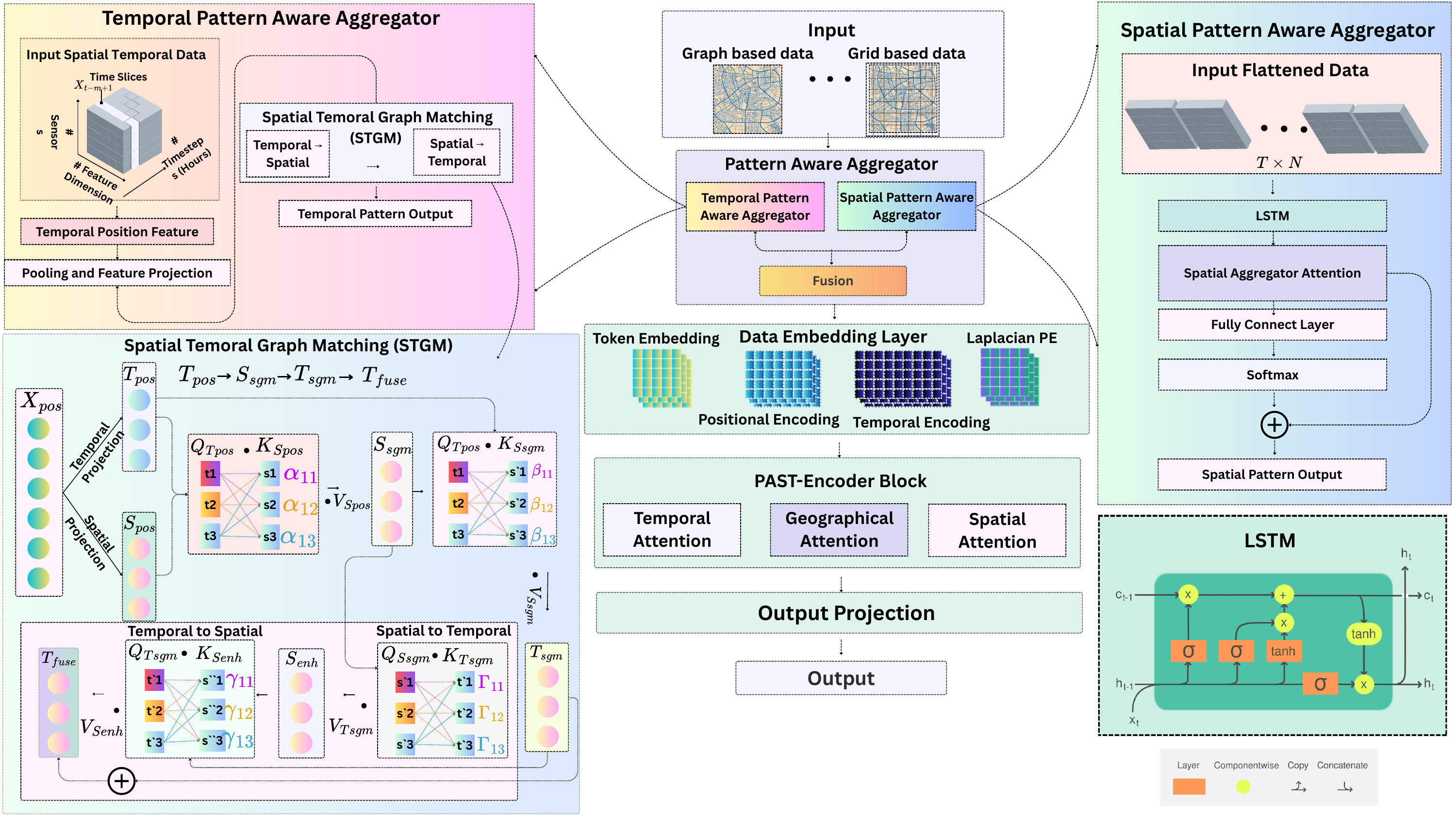}
    \caption{The framework of STPFormer}
    \label{fig:framework}
\end{figure}

As shown in Figure~\ref{fig:framework}, STPFormer consists of four key modules that cooperate to model intricate spatiotemporal patterns:
(1) a Data Embedding Layer that encodes raw traffic data along with spatial identifiers and temporal indicators;
(2) a Pattern-Aware Block, which includes the Temporal Position Aggregator (TPA) and the Spatial Sequence Aggregator (SSA), designed to capture long-range temporal dependencies and dynamic spatial interactions;
(3) a Spatial–Temporal Transformer Block, which acts as an Attention Mixer to hierarchically fuse encoder outputs across layers and modalities;
(4) an Output Projection Layer that transforms the aggregated features into final traffic predictions.

In the TPA module, we propose a dedicated spatio-temporal Graph Matching (STGM) model to refine temporal representations through bidirectional spatiotemporal alignment. As show in the bottom-left of Figure~\ref{fig:framework}, each temporal token $t$ and spatial token $s$ is derived via learnable linear projections from the position-enhanced input $\mathbf{X}_{\text{pos}}$.  
Intermediate representations $t'$, $s'$, and $s''$ correspond to attention-refined features after each stage of message exchange.  
The four attention maps—$\alpha$, $\beta$, $\gamma$, and $\delta$—denote the learned similarity weights for each directional alignment, respectively, and correspond to the formulations in Equations~(~\eqref{eq:att2} –\eqref{eq:att5}).  
This design enables STGM to model fine-grained element-wise correlations and effectively capture long-range dependencies across spatial and temporal domains.

\subsection{Pattern Aware Block}

\subsubsection{Spatial–Sequence Aggregator (SSA)}
Traditional graph convolution operations process nodes in a pairwise manner, which often fails to capture broader, corridor-level correlations that span multiple network hops. These long-range spatial dependencies are crucial for understanding traffic flow patterns that extend across entire transportation corridors or ring roads.

To address this limitation, we introduce the Spatial–Sequence Aggregator (SSA), which captures long-range spatial dependencies through a sequential processing approach. Given the spatiotemporal input $\mathbf{X}:=\mathbf{X}_{t-m+1:t} \in \mathbb{R}^{m \times N \times d}$ over the past $m$ time steps, the SSA linearizes each $m \times N$ traffic frame into a token sequence:
\begin{equation}
    \mathbf{X}_{\text{seq}} = \text{Reshape}_{(mN, d)}(\mathbf{X} ),
\end{equation}
where $\mathbf{X}_{\text{seq}} \in \mathbb{R}^{(m \cdot N) \times d}$ is the flattened spatiotemporal representation treated as a sequence of $m \cdot N$ tokens.
The sequential representation is then processed using a combination of LSTM and multi-head attention mechanisms:
\begin{equation}
    \mathbf{H}_{\text{SSA}} = \text{LSTM}(\mathbf{X}_{\text{seq}}) + \text{MHAttn}(\mathbf{X}_{\text{seq}}).
\end{equation}
Here, $\text{LSTM}(\cdot)$ captures sequential dependencies using a standard long short-term memory unit, while $\text{MHAttn}(\cdot)$ applies a multi-head self-attention mechanism to model global interactions across the token sequence. The two outputs are combined via residual addition to obtain the intermediate representation $\mathbf{H}_{\text{SSA}} \in \mathbb{R}^{(m \cdot N) \times d_h}$.

A residual gating mechanism is applied to preserve fine-grained information from the spatiotemporal representation output by the spatial sequence aggregator. Specifically, we define the gated output as:
\begin{equation}
    \mathbf{S}_{\text{seq}} = \sigma(\mathbf{W}_g \mathbf{H}_{\text{SSA}}) \odot \mathbf{H}_{\text{SSA}},
\end{equation}
where $\mathbf{H}_{\text{SSA}} \in \mathbb{R}^{(m \cdot N) \times d_h}$ is the hidden representation obtained after sequential modeling, $\mathbf{W}_g \in \mathbb{R}^{d_h \times d_h}$ is a shared learnable weight matrix applied to each spatiotemporal token, and $\sigma(\cdot)$ denotes the softmax activation function (as used in our implementation). The gating mechanism $\sigma(\mathbf{W}_g \mathbf{H}_{\text{SSA}})$ controls the flow of information through element-wise multiplication ($\odot$), enabling the network to selectively retain important features from the original representation $\mathbf{H}_{\text{SSA}}$.

Finally, the processed sequence is reshaped back to the original tensor format:
$\mathbf{S} = \text{Reshape}_{(m, N, d_h)}(\mathbf{S}_{\text{seq}})$. The output $\mathbf{S}$ will later be integrated into the Attention Mixer section, where it contributes to the hierarchical aggregation of encoder features, allowing the model to retain and enhance spatial representations across all layers.
This design enables the SSA to learn comprehensive spatial patterns such as tidal flows along bus corridors or ring-road traffic dynamics that would be difficult to recover using purely local convolution kernels.

\subsubsection{Temporal–Position Aggregator (TPA)}
Temporal dynamics in traffic systems often exhibit position-sensitive patterns, where certain time steps (e.g., rush hours) carry stronger temporal signals than others. To effectively incorporate such temporal priors, we introduce the Temporal–Position Aggregator (TPA), which enhances temporal representations through positional encoding and structure-aware alignment.

Given the spatiotemporal input $\mathbf{X} \in \mathbb{R}^{m \times N \times d}$ over the past $m$ time steps, we first compress the spatial dimension using average pooling to obtain a temporal sequence. Learnable positional priors are then added:
\begin{equation}
    \mathbf{X}_{\text{pos}} = \text{Pool}(\mathbf{X}) + \text{PosEmb}(1:m).
\end{equation}
Here, $\text{Pool}(\cdot)$ denotes spatial average pooling across $N$ nodes, and $\text{PosEmb}(1:m)$ injects learnable random walk-based temporal position embeddings for each time step $1$ to $m$.

The resulting temporal sequence $\mathbf{X}_{\text{pos}} \in \mathbb{R}^{m \times d}$ is then passed through a two-layer feed-forward network (FFN), which maps the input dimension $d$ to a hidden dimension $d_h$, to transform temporal features. These features are then processed by the Spatial-Temporal Graph Matching module (STGM) for fine-grained alignment:
\begin{equation}
    \mathbf{H}_{\text{TPA}} = \text{STGM}(\text{FFN}(\mathbf{X}_{\text{pos}})).
    \label{eq:htpa}
\end{equation}

The output $\mathbf{H}_{\text{TPA}}$ is linearly projected and spatially expanded to form a tensor $\mathbf{T} = \text{Expand}(\mathbf{H}_{\text{TPA}}) \in \mathbb{R}^{h \times N \times d_h}$, restoring the target spatiotemporal shape for prediction. The result $\mathbf{T}$ will later be integrated with result $\mathbf{S}$ of SSA into the Attention Mixer section, where it contributes to attention computation by injecting temporally aligned signals.

This design allows TPA to distinguish position-sensitive time steps and infuse temporal structure into the attention backbone. The core of this temporal alignment is handled by the STGM module, detailed in the following section

\subsubsection{Spatial-Temporal Graph Matching Module (STGM)}

To refine temporal representations with spatial context, we introduce the Spatio-Temporal Graph Matching Module (STGM) as a sub-module within the Temporal Position Aggregator (TPA), which performs bi-directional attention alignment between spatial and temporal projections of the pooled input.

Specifically, the input to STGM is the temporally pooled and position-enhanced sequence $\mathbf{X}_{\text{pos}} \in \mathbb{R}^{m \times d_h}$, produced within the Temporal Position Aggregator (TPA). Two learnable linear projections are applied to $\mathbf{X}_{\text{pos}}$ to obtain the temporal and spatial input branches of the module:
\begin{equation}
\mathbf{T}_{\text{pos}} = \mathbf{X}_{\text{pos}} \mathbf{W}_t, \quad \mathbf{S}_{\text{pos}} = \mathbf{X}_{\text{pos}} \mathbf{W}_s,
\label{eq:att1}
\end{equation}
where $\mathbf{W}_t, \mathbf{W}_s \in \mathbb{R}^{d_h \times d_h}$ are learnable projection matrices.

First, we compute attention from $\mathbf{T}_{\text{pos}}$ to $\mathbf{S}_{\text{pos}}$ to extract temporally guided spatial features. The resulting attention output is concatenated with $\mathbf{S}_{\text{pos}}$ and linearly projected to obtain the enhanced spatial representation:
\begin{equation}
\mathbf{S}_{\text{sgm}} = [\mathbf{S}_{\text{pos}};\ \text{Attn}_{T \rightarrow S}(\mathbf{T}_{\text{pos}})] \cdot \mathbf{W}_H,
\label{eq:att2}
\end{equation}
where $\mathbf{W}_H \in \mathbb{R}^{2d_h \times d_h}$ is a learnable parameter. The factor $2$ arises because the enhanced spatial representation is formed by concatenating the original spatial projection and the temporally guided attention output, effectively doubling the feature dimension before projection back to $d_h$.

Next, the enhanced spatial features $\mathbf{S}_{\text{sgm}}$ are used to guide temporal representation refinement:
\begin{equation}
\mathbf{T}_{\text{sgm}} = \text{Attn}_{S \rightarrow T}(\mathbf{S}_{\text{sgm}}) + \mathbf{T}_{\text{pos}}.
\label{eq:att3}
\end{equation}

Here, $\mathbf{S}_{\text{sgm}}, \mathbf{T}_{\text{sgm}} \in \mathbb{R}^{m \times d_h}$ denote the spatially and temporally enriched representations, respectively.

\paragraph{Temporal-to-Spatial Attention.} 
In the first stage, spatial tokens absorb contextual information from temporal tokens via dot-product attention. Here, $\mathbf{S}_{\text{sgm}}$ serves as queries, and $\mathbf{T}_{\text{sgm}}$ as keys and values:
\begin{equation}
\mathbf{S}_{\text{enh}} = \text{Softmax}(\mathbf{S}_{\text{sgm}} \mathbf{T}_{\text{sgm}}^\top) \mathbf{T}_{\text{sgm}}.
\label{eq:att4}
\end{equation}
This implements a single-head attention mechanism, where the softmax is applied row-wise across the temporal dimension. The attention map determines how each spatial token attends to temporal cues for context-aware enhancement.

\paragraph{Spatial-to-Temporal Attention.}
In the second stage, the enhanced spatial features act as a guide to refine temporal representations:
\begin{equation}
\mathbf{T}_{\text{fused}} = \text{Softmax}(\mathbf{T}_{\text{sgm}} \mathbf{S}_{\text{enh}}^\top) \mathbf{S}_{\text{enh}} + \mathbf{T}_{\text{sgm}}.
\label{eq:att5}
\end{equation}
This reversed attention uses $\mathbf{T}_{\text{sgm}}$ as queries and $\mathbf{S}_{\text{enh}}$ as keys and values, producing a spatially guided temporal output. A residual connection is added to preserve the original temporal signal and improve model stability.

The final output $\mathbf{T}_{\text{fused}} \in \mathbb{R}^{m \times d_h}$ is broadcasted to shape $\mathbb{R}^{m \times N \times d_h}$ and serves as the temporally aligned bias for the Temporal–Position Aggregator (TPA). This dual attention mechanism allows spatial context to guide the refinement of temporal structures, enhancing the modeling of fine-grained traffic dynamics. The final output $\mathbf{T}_{\text{fused}}$ is part of TPA and will be returned according to Equation~\eqref{eq:htpa}. To be consistent, we write the output $\mathbf{T}_{\text{fused}}$ as $\mathbf{H}_{\text{TPA}}$.

\subsection{Data Embedding Layer}
\label{subsec:data_embedding}

In the Spatial Graph Embedding~\cite{10.1609/aaai.v37i4.25556}, spectral graph theory is used to generate a global structure-aware embedding for each time step $t$, formulated as $\mathbf{X}_{t,\text{spe}} = \mathbf{U}_{\text{spe}} \mathbf{W}_{\text{spe}} + \mathbf{b}_{\text{spe}}$, where $\mathbf{U}_{\text{spe}}$ consists of the top-$k$ eigenvectors of the Laplacian matrix, capturing the road network topology. For the Temporal Periodic Embedding, periodic traffic cycles are encoded by decomposing each timestamp $t$ into a weekly index $w(t)$ and a daily index $d(t)$, which are embedded as $\mathbf{X}_w, \mathbf{X}_d \in \mathbb{R}^{m \times d}$ to capture multi-scale temporal patterns consistent with the spatial features~\cite{10.1609/aaai.v37i4.25556}. For the Temporal Positional Encoding, the absolute position of each timestamp is represented using a sinusoidal encoding~\cite{VAS}: $\mathbf{X}_{t,\text{tpe}}(i) = \sin(\frac{t}{10000^{2i/d}})$ for even $i$ and $\cos(\frac{t}{10000^{2(i-1)/d}})$ for odd $i$, forming the matrix $\mathbf{X}_{\text{tpe}} \in \mathbb{R}^{m \times d}$, which provides continuous position-aware information. Finally, the Embedding Integration step combines all components through element-wise summation with appropriate broadcasting. The unified embedding $\mathbf{X}_{\text{emb}} \in \mathbb{R}^{m \times N \times d}$ is computed as:
\begin{equation}
\mathbf{X}_{\text{emb}} = \mathbf{X}_{\text{data}} + \mathbf{X}_{\text{spe}, t} + \mathbf{X}_{w,n} + \mathbf{X}_{d,n} + \mathbf{X}_{\text{tpe},n},
\label{eq:embedding_integration}
\end{equation}
where $\mathbf{X}_{\text{spe}, t} \in \mathbb{R}^{N \times d}$ is broadcast along the temporal axis, while $\mathbf{X}_{w,n}, \mathbf{X}_{d,n}, \mathbf{X}_{\text{tpe},n} \in \mathbb{R}^{m \times d}$ are broadcast across spatial nodes. For clarity, we denote $\mathbf{X} \equiv \mathbf{X}_{\text{emb}}$ hereafter.

\subsection{Pattern Aware Spatial Temporal Encoder Block (PAST-Encoder Block)}\label{sec:PAST-Encoder Block}
The Pattern-Aware Spatio-Temporal Encoder comprises three attention blocks: temporal attention, geographical attention, and spatial attention, which process the inputs $\mathbf{X}_t$ and $\mathbf{X}_n$ 
, respectively. This design follows the approach proposed in~\cite{10.1609/aaai.v37i4.25556}. An example of the spatial attention mechanism at each time step $t$ is illustrated as follows:
$\mathbf{Q}_t = \mathbf{X}_{t} \mathbf{W}_Q$, 
$\mathbf{K}_t = \mathbf{X}_{t} \mathbf{W}_K$, and 
$\mathbf{V}_t = \mathbf{X}_{t} \mathbf{W}_V$, 
where $\mathbf{W}_Q, \mathbf{W}_K, \mathbf{W}_V \in \mathbb{R}^{d \times d_0}$ are learnable matrices and $d_0$ is the hidden dimension.  
Spatial, geometrical and temporal self-attention are then computed as the following respectively:
\begingroup            
\small                 
\begin{align}
  \mathbf{X}_{t}^{\text{spat}} &= 
    \text{Softmax}\!\bigl(\tfrac{\mathbf{Q}_t\mathbf{K}_t^{\top}}{\sqrt{d_0}}\bigr)\,\mathbf{M}_{\text{spat}}, \\[2pt]
  \mathbf{X}_{t}^{\text{geo}} \label{M}&=
    \text{Softmax}\!\bigl(\tfrac{\mathbf{Q}_t\mathbf{K}_t^{\top}}{\sqrt{d_0}}\bigr)\,\mathbf{M}_{\text{geo}}, \\[2pt]
  \mathbf{X}_{n}^{\text{temp}} &=
    \text{Softmax}\!\bigl(\tfrac{\mathbf{Q}_n\mathbf{K}_n^{\top}}{\sqrt{d_0}}\bigr).
\end{align}
\endgroup

Here, $\mathbf{M}_{\text{spat}}$ and $\mathbf{M}_{\text{geo}}$ are binary spatial and geographical masking matrices, respectively, which assign a weight of 1 to the connections between a node and its spatial or geographical neighbors, and 0 otherwise.  
The outputs from the geo-heads, spatial-heads, and temporal-heads of the multi-head attention are concatenated and then projected through a shared linear transformation:
\begin{equation}
\text{PAST-Encoder} \!= \!\bigoplus \!\left( \!\mathbf{X}^{\text{geo}}_{1,\ldots,h_{\text{geo}}},\; \mathbf{X}^{\text{spat}}_{1,\ldots,h_{\text{spat}}},\; \mathbf{X}^{\text{temp}}_{1,\ldots,h_{\text{temp}}} \!\right)\! \mathbf{W}^{O},
\label{4}
\end{equation}
where $\bigoplus$ denotes the concatenation along the head dimension, $h_{\text{geo}}, h_{\text{spat}}, h_{\text{temp}}$ are the number of heads for geographical, spatial, and temporal attention respectively, and $\mathbf{W}^{O} \in \mathbb{R}^{d \times d}$ is a learnable output projection matrix.  
Finally, the output of this multi-head self-attention is fed into a position-wise feed-forward network to obtain the updated representation $\mathbf{H}^{(l)} \in \mathbb{R}^{m \times N \times d}$.

\subsection{Attention   Mixer and Output Layer}

\paragraph{Attention Mixer}
The Attention Mixer integrates multi-perspective spatio-temporal features and performs hierarchical attention-based refinement. It begins by constructing an enhanced embedding $\mathbf{X}_{\text{mix}}$ through the element-wise addition of the raw input $\mathbf{X}$ embedding, the temporal-aware bias $\mathbf{T}$ from the Temporal Position Aggregator (TPA), and the spatial-aware representation $\mathbf{S}$ from the Spatial Sequence Aggregator (SSA):
\begin{equation}
\mathbf{X}^{\text{mix}} = \text{Embed}(\mathbf{X}) + \mathbf{T} + \mathbf{S}.
\end{equation}
This fused representation $\mathbf{X}_{\text{mix}} \in \mathbb{R}^{m \times N \times d_h}$ is then processed by a stack of $L$ encoder blocks, each designed to capture complex temporal and spatial patterns. Specifically, $\mathbf{X}^{\text{mix}}$ is split into temporal $\mathbf{X}^{\text{mix}}_n$ and spatial $\mathbf{X}^{\text{mix}}_t$ and then feed into the PAST-EncoderBlock, and correspond to the formulations in Equations~\eqref{M}–-\eqref{4}. In here, each encoder block incorporates attention-based modules—including multi-head self-attention and pattern-aware matching—to refine spatio-temporal representations:
\begin{equation}
\mathbf{H}^{(l)} = \text{PAST-EncoderBlock}^{(l)}(\mathbf{X}_{\text{mix}}), \quad l = 1, \dots, L.
\end{equation}
To align feature formats across all encoder layers, the outputs $\mathbf{H}^{(l)} \in \mathbb{R}^{m \times N \times d_h}$ are permuted and projected via learnable $1 \times 1$ convolutions:
\begin{equation}
\mathbf{M}^{(l)} = \mathrm{Conv}_{1\times1}^{(l)}(\mathrm{Permute}(\mathbf{H}^{(l)})).
\end{equation}
Here, $\mathbf{M}^{(l)} \in \mathbb{R}^{d_h \times N \times m}$. The projected multi-level outputs are then aggregated by residual summation to obtain the final mixed spatio-temporal feature representation:
\begin{equation}
\mathbf{S}_{\text{mix}} = \sum_{l=1}^{L} \mathbf{M}^{(l)}.
\end{equation}
By combining fused embeddings with multiple attention-enhanced encoder layers, the Attention Mixer facilitates deep integration of spatial and temporal patterns across multiple levels. The resulting tensor $\mathbf{S}_{\text{mix}}$ is passed into the output module for traffic forecasting.

\paragraph{Output Layer}
The output layer transforms the aggregated spatio-temporal feature tensor $\mathbf{S}_{\text{mix}} \in \mathbb{R}^{d_h \times N \times m}$ into the final prediction. It first applies a point-wise $1 \times 1$ convolution to project the channel dimension to the output feature size $d_{\text{out}}$, followed by a second convolution to reshape the temporal axis into the forecasting horizon $h$:
\begin{equation}
\widehat{\mathbf{Y}} = \mathrm{Conv}_{\text{out}}\left( \mathrm{ReLU}\left(\mathrm{Conv}_{1\times1}(\mathbf{S}_{\text{mix}})\right) \right).
\end{equation}
Here, $\widehat{\mathbf{Y}} \in \mathbb{R}^{d_{\text{out}} \times N \times h}$ is the predicted traffic state over the forecast horizon. This design preserves spatial structure while enabling flexible output generation across various prediction tasks.

\section{Experiment}
\label{experiment}
We evaluate our model on five benchmark traffic datasets, PeMS04, PeMS07, PeMS08, NYCTaxi, and CHIBike, spanning diverse spatio-temporal patterns and accessible via the LibCity repository~\cite{wang2023libcity}.

\subsection{Datasets and Experiment Setup}
\begin{table}[htbp]
\centering
\caption{Dataset Information}
\normalsize  
\resizebox{0.5\textwidth}{!}{  
\begin{tabular}{|c|c|c|c|c|c|}
\hline
Dataset & Nodes & Edges & Seps & Interval & Time Range \\
\hline
PeMS04 & 307 & 340 & 16992 & 5 min & 01/01/2018 -- 02/28/2018 \\
PeMS07 & 883 & 866 & 28224 & 5 min & 05/01/2017 -- 08/31/2017 \\
PeMS08 & 170 & 295 & 17856 & 5 min & 07/01/2016 -- 08/31/2016 \\
NYCTaxi & 75 (15×5) & 484 & 17520 & 30 min & 01/01/2014 -- 12/31/2014 \\
CHIBike & 270 (15×18) & 1966 & 4416 & 30 min & 07/01/2020 -- 09/30/2020 \\
\hline
\end{tabular}
}
\label{tab:datasets_half}
\end{table}

\paragraph{Datasets} The freeway datasets PeMS04, PeMS07, and PeMS08 capture 5-minute traffic volumes collected by Caltrans loop detectors; each sensor is represented as a graph node and edges follow physical road connectivity, with PeMS07 being the most extensive (883 nodes, 866 edges) \cite{song2020spatial}. NYCTaxi provides 30-minute taxi demand on a \(15\times5\) Manhattan grid but lacks an explicit spatial graph, requiring models to infer latent spatial dependencies \cite{liu2020dynamic}. CHIBike contains 30-minute rental counts from 270 bike stations projected onto a \(15\times18\) grid and features sparse, irregular, and spatially imbalanced demand \cite{VAS}.


\paragraph{Experiment Setup}Training is performed on a virtualized GPU (vGPU 48GB). We set the batch size to 16 across all experiments. The model is trained using the AdamW optimizer with an initial learning rate of 0.001, a cosine learning rate scheduler, and a 5-epoch warm-up phase. Early stopping is applied based on validation loss with a patience of 50 epochs. A fixed random seed (seed=1) is set for Python, NumPy, and PyTorch to ensure the reproducibility of results. All results are reported based on a single run unless otherwise stated.

\subsection{Experimental Results and Analysis}

\begin{table}[htbp]
\caption{Performance on state-based Datasets.}
    \label{tab:performance}
\hspace{-0.2cm}\includegraphics[width=1.1\linewidth]{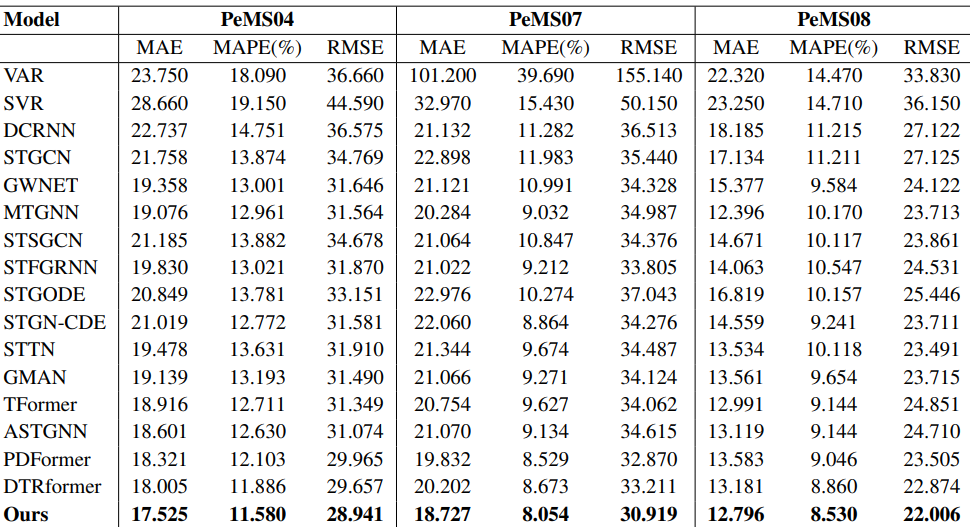}
\end{table}

We compare STPFormer (ours) against diverse baselines on both sensor-based (PeMS04/07/08) and grid-based (NYCTaxi, CHIBike) datasets, using MAE, MAPE, and RMSE as metrics (Table~\ref{tab:performance}, Table~\ref{tab:2}). Key observations are as follows:

(1) The statistical models (VAR, SVR) lack clear spatial structure modeling, so their performance is relatively poor, especially on PeMS08. However, the data of STPFormer(our) is excellent, being 12.796.

(2) The classic GNN-based methods (GWNET, STGCN, STSGCN) assume static topological structures. While STPFormer (our) dynamically models spatial dependence, compared with STGCN, the MAE of NYCTaxi inflow is reduced by 33.7\%, and it achieves lower errors than the recent sota model DTRFormer.

(3) Advanced GNNS (MTGNN, STGNCDE) perform poorly in irregular grids. STPFormer (ours) has achieved stronger generalization than traditional gnn and DTRFormer.

(4) Attention-based models (ASTGNN, STTN) may overfit sparse regions. STPFormer (our) enhances long-term dependency capture and achieves the best MAE and RMSE on CHIBike inflows, which are 0.365 and 1.381 respectively.

(5) Models based on cnn (STResNet, DSAN, DMSSTNet) usually do not have time alignment and remote memory designed. STPFormer (our) demonstrates the ability of sota in both dense grids and sparse sensor networks.

\begin{table}[htbp]
\caption{Performance on state-based Datasets..}
    \label{tab:2}
\hspace{-0.5cm}\includegraphics[width=1.15\linewidth]{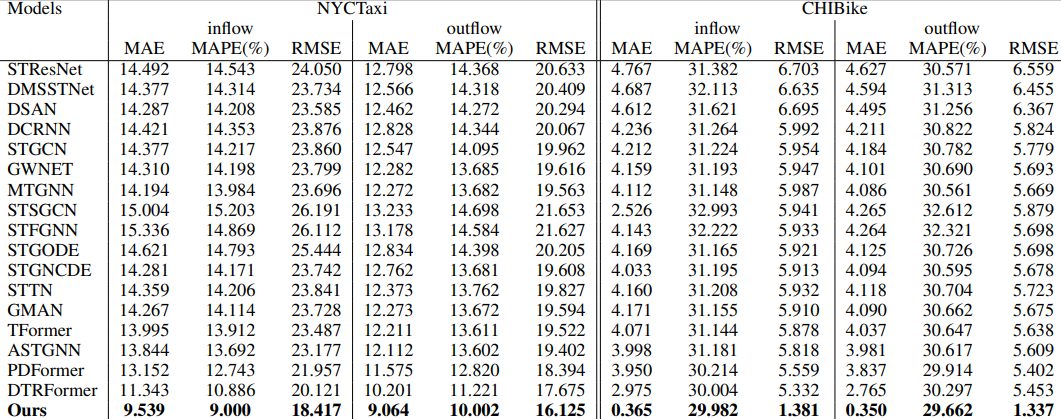}  
\end{table}

\subsection{Ablation Study}
\begin{figure}[htbp]
    \centering
    \includegraphics[width=0.7\linewidth]{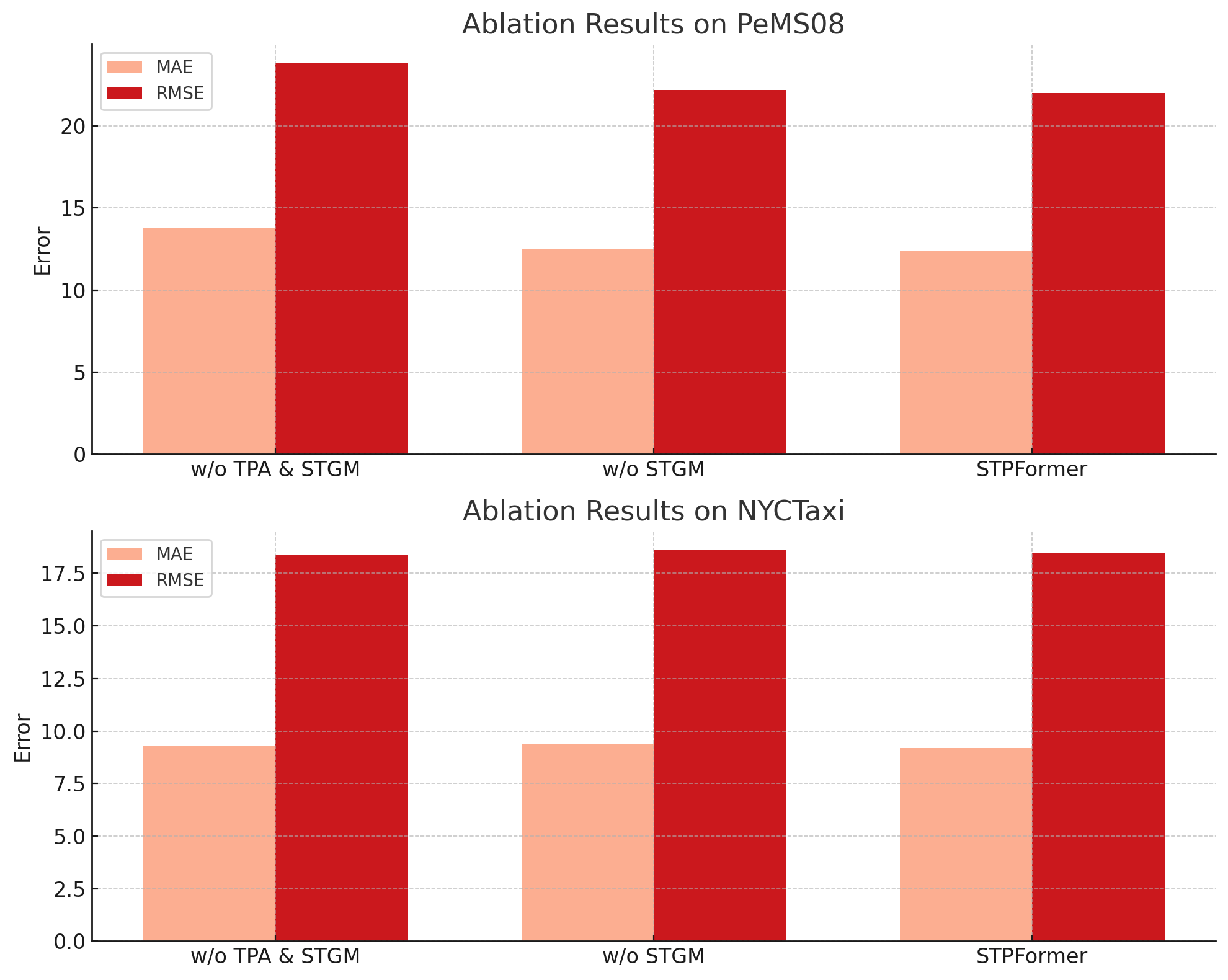}
    \caption{Ablation study on PeMS08 and NYCTaxi.}
    \label{fig:ablation}
\end{figure}

We conduct ablation studies on PeMS08 and NYCTaxi to evaluate three forms: (i) w/o TPA \& STGM, (ii) w/o STGM, and (iii) the full STPFormer (ours). Figure~\ref{fig:ablation} shows that removing TPA and STGM will lead to very high errors. Adding TPA will improve this result, and the complete STPFormer (our) shows the performance of sota on both datasets.

\begin{figure}[htbp]
\hspace{-0.6cm}\includegraphics[width=1.14\linewidth]{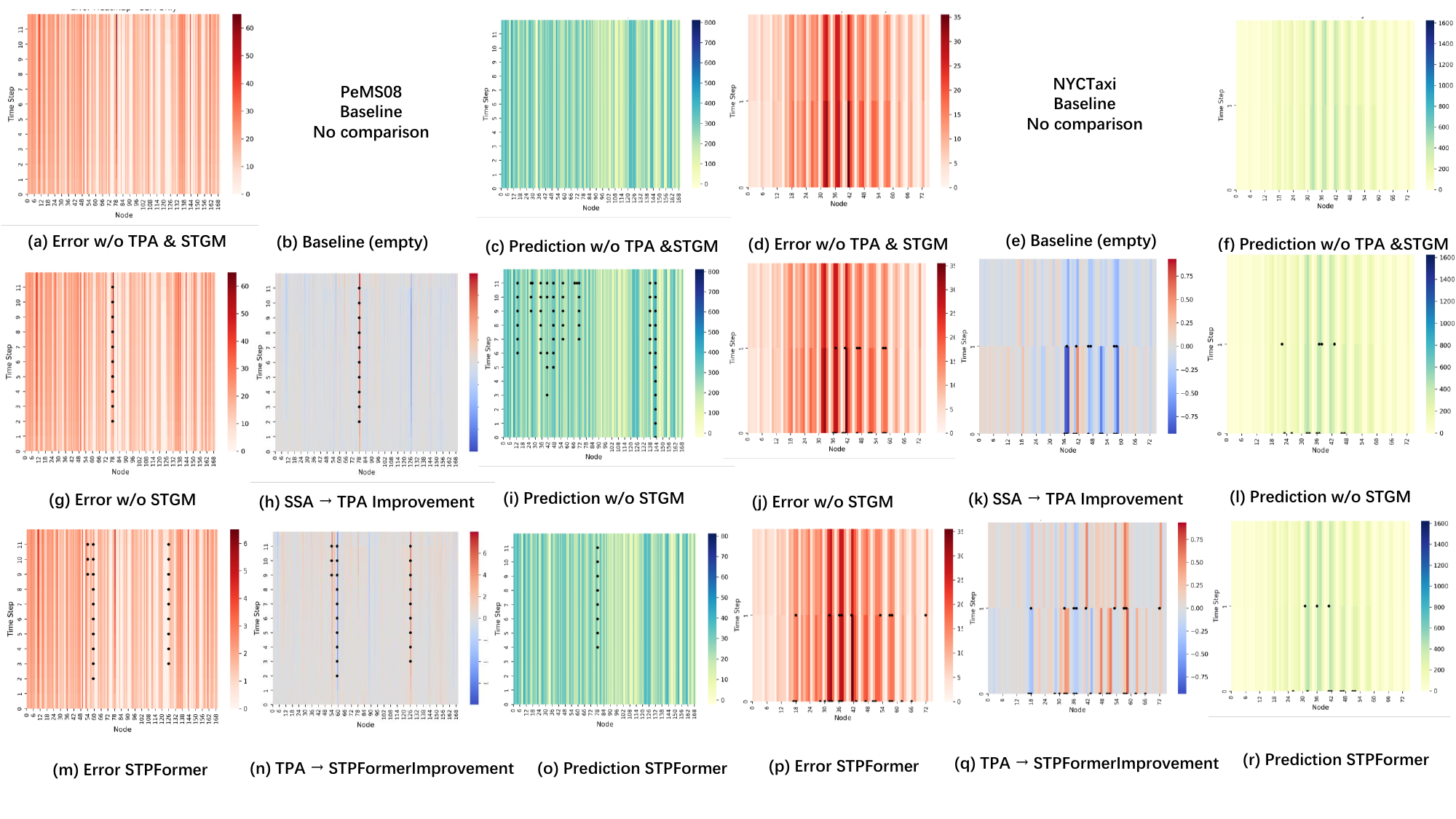}
    \caption{Ablation of model on PeMS08 and NYCTaxi.}
    \label{fig:ablation-pems08-taxi}
\end{figure}
\paragraph{PeMS08 (State-based).}
This dataset has strong periodicity and spatial heterogeneity. As shown in Figure~\ref{fig:ablation-pems08-taxi}(a), (w/o TPA \& STGM) can lead to relatively high spatiotemporal errors, especially during peak periods, such as the error between 60 and 100 nodes. Its output (c) differs significantly from the daily cycle, indicating that modeling time dependence using only the Spatial Sequence Aggregator (SSA) is not ideal.
Using the time-location aggregator (w/o STGM), the error decreased significantly in the high-variance region (g), and time alignment improved the predictions with time steps ranging from 4 to 10 and nodes ranging from 60 to 100 (h). The output (i) of (w/o STGM) is more stable than that of (w/o TPA \& STGM). Adding the Spatial Temporal Graph Matching module (STGM) further enhances the performance of the model: the errors of edge nodes are significantly reduced, such as nodes 12 and 288, graph (m). The final output (o) of STPFormer (ours) shows superior spatiotemporal consistency, verifying the complementary role of all integrated modules.

\paragraph{NYCTaxi (Grid-based).}
Unlike PeMS08, the characteristics of NYCTaxi are high spatial variability and weak temporal periodicity. As shown in Figure \ref{fig:ablation-pems08-taxi} (d), the form of (w/o TPA \& STGM) presents dispersed high-error regions, especially in the central and lower grid positions, such as 30 to 50 and 60 to 75. The output (f) is unstable in time and inconsistent in space, which indicates that it is not ideal to use the Spatial Sequence Aggregator (SSA) alone to process grid structures and irregular data. After adding the Temporal Position Aggregator (TPA) (j), there is a significant improvement in the middle grid area.
The improved heatmap (k) shows that the improvement is local and sparse, indicating that the temporal consistency remains unsatisfactory. In contrast, the form (l) without the spatio-temporal graph matching module shows a more organized structure, but there are still some deficiencies. After merging the spatio-temporal graph matching module, STPFormer achieved greater improvement: The error graph (p) became smoother, and the errors in areas 25 to 45 and 66 to 78 were significantly reduced. Furthermore, the improved graph (q) demonstrates that integrating spatial memory can significantly enhance the model. The final prediction output (r) of STPFormer (ours) maintains a high degree of spatiotemporal consistency and captures the urban taxi traffic patterns more accurately.

\subsection{Visual of Data Transformations Across Modules}
\begin{figure}[htbp]
    \centering
    \includegraphics[width=1.05\linewidth]{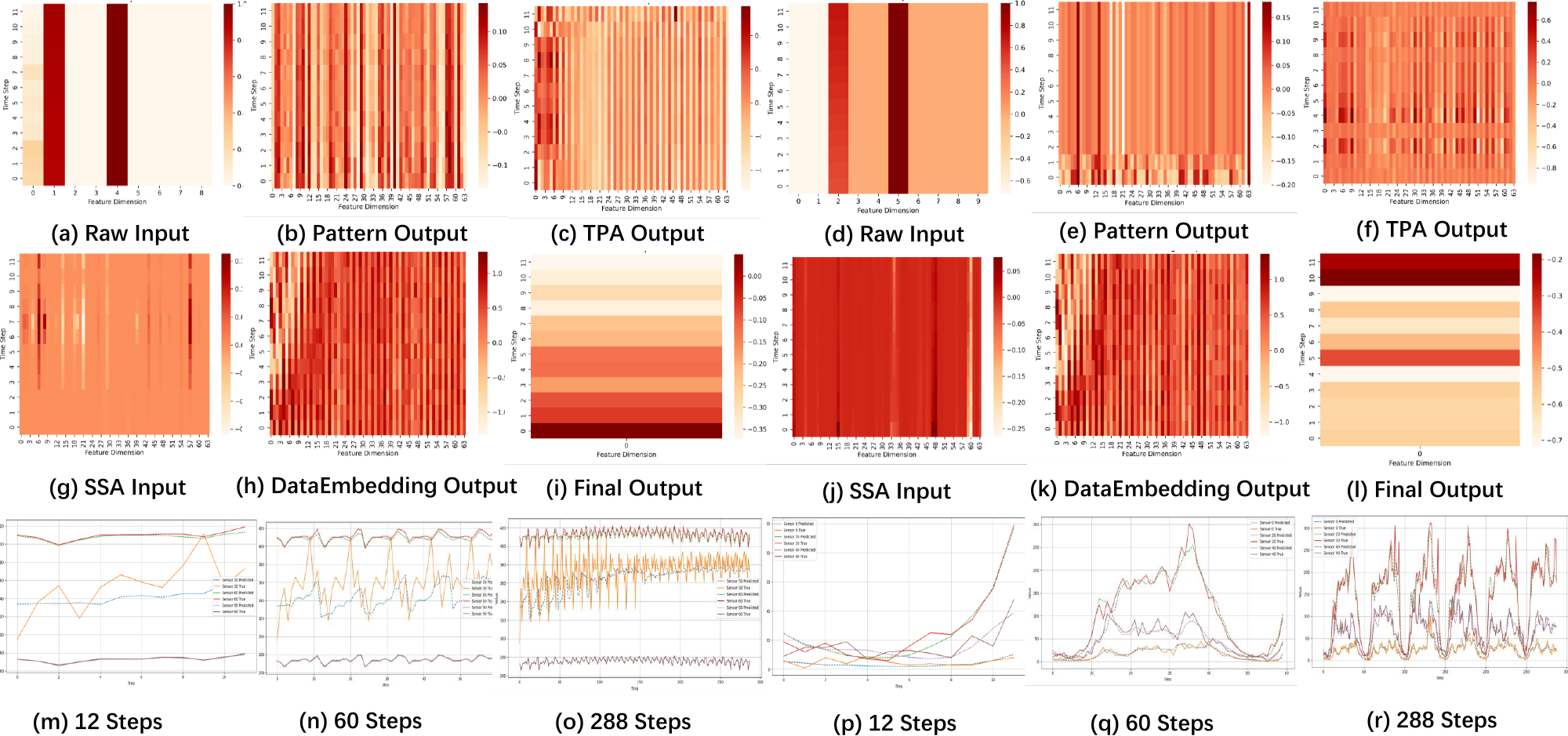}
    \caption{Visualization of module output and time steps.}
    \label{6}
\end{figure}

To analyze the contributions of each module, we visualize the intermediate output of PeMS08 and NYCTaxi. These two datasets are different in spatial structure. STPFormer (ours) is always learning meaningful spatiotemporal patterns between the two.

\paragraph{PeMS08 (State-based).}
As shown in Figure \ref{6}, the original input (a) is sparse and fragmented. After using the Pattern module (b), the vertical time pattern becomes clearer and the signal distribution is more uniform, which means that the Pattern module can better capture repetitive trends. Temporal Position (TPA) (c) helps keep the time series more consistent by connecting points over a longer period of time. For the Spatial Sequence Aggregator (SSA) input (g), we can observe the process of adding local spatial information. Although the data layout is sparse, the important links between the data are retained. Data Embedding output (h) adds additional information, such as the time of day, making the signal denser and smoother over time. Finally, the results of STPFormer (our) (i) show clear and well-organized patterns, and become smoother over time.
The prediction results (m-o) verify the robustness of STPFormer (our) at all levels. The predictions for the short term (12 hours and 60 steps) and the long term (288 steps) are both very close to the real data, proving that the model has a very strong generalization ability in the daily traffic.

\paragraph{NYCTaxi (Grid-based).}
As shown in Figure \ref{6}, the original input (d) indicates that most of the features are sparse and of low variation, which means that the appearance of taxis is very uneven and concentrated in certain areas. Unlike the PeMS08 data, NYCTaxi does not have a clear road network structure, but there are strong regional differences. For instance, the downtown area is very active, while many other places have almost no demand. The Spatial Sequence Aggregator (SSA) (j) learns spatial dependencies without an explicit adjacency graph, identifying correlated grids and filtering noise in inactive areas. The Data Embedding (k) injects position-aware features, helping distinguish spatially similar but functionally distinct grids, like similar volume but different peak hours. The final output (l) shows the predicted taxi demand for each future time step. This illustrates how STPFormer (ours) turns the complex spatio-temporal features it learns into clear and understandable demand. The Prediction Curve (p-r) clearly indicates that for short-term predictions, the model closely follows the actual trend, while for longer-term predictions, the model still maintains a smooth transition under sudden changes. This proves that even when the requirements change greatly and there is no clear spatial structure, this model can still work well.

Combined with the results of PeMS08, STPFormer (ours) can transform the original data into clear and consistent spatiotemporal patterns. For PeMS08, TPA helps to better align the time series, and SSA makes the structure smoother. Take NYCTaxi as an example. The results show that STPFormer (ours) can effectively handle different types of traffic data and has a strong generalization ability

\section{Related Work}
Traffic forecasting is challenging because it must capture complex patterns in both space and time, often under irregular and dynamic conditions. Traditional statistical methods, such as ARIMA~\cite{Zhou2006} and the Vector AutoRegressive model~\cite{Luetkepohl2005}, rely on assumptions of linearity and stationarity, which limit their flexibility in practice. Deep learning models, including CNNs~\cite{NIPS2012_c399862d} and RNNs~\cite{6894591}, can learn nonlinear relationships but often treat spatial and temporal features separately, but they dealt with it independently.

Graph Neural Networks (GNNs) improve spatial modeling by leveraging topological information. Models like DCRNN~\cite{LiYS018} and STGCN~\cite{Zhang2016} have shown promising results but depend on static graphs and fixed temporal windows, which can restrict adaptability. Later models, such as MTGNN~\cite{wu2020} and DGCRN~\cite{10.1145/3532611}, use adaptive graphs and dynamic connections, yet they often still face challenges in fully modeling temporal patterns and smoothly integrating spatial and temporal contexts. Attempts like DFormer~\cite{zhao2024} aim to couple spatial and temporal components more tightly.

The transformer structure has indeed performed outstandingly in practice due to its own excellent characteristics, such as the global acceptance domain and modeling flexibility, such as GMAN~\cite{Zheng2020}, STTN~\cite{s24175502}, and AutoST~\cite{10.1145/3442381.3449816} adapt vanilla Transformers to spatiotemporal data but  it has also been criticized by researchers for using static spatial graphs or fixed-position encoding. Recent methods like STFormer~\cite{Qin2021} and DGFormer~\cite{Xu2024} introduce adaptive attention on learned graph structures but remain tailored to either state-based or grid-based data. Hybrid approaches, including PDFormer~\cite{10.1609/aaai.v37i4.25556},  CCDSReFormer \cite{ccdsreformertrafficflowprediction} and ST-LLMDF \cite{2024stllmdf} improve flexibility but can struggle with interpretability or input heterogeneity.
  
Recently, selective state space models have emerged as a powerful alternative for spatiotemporal forecasting due to their linear complexity and ability to handle long-range dependencies. Models such as ST-Mamba~\cite{shao2024stmamba}, STDAtt-Mamba \cite{2123-33604}, and ST-MambaSync~\cite{STMambaSync} integrate selective state space dynamics with spatial attention mechanisms to capture both local and global structures. These architectures enhance interpretability while maintaining competitive accuracy but lacks an inherent mechanism to model spatial dependencies, which are essential for accurately capturing the localized interactions between nodes. 

To address these gaps, we propose a unified architecture that integrates multi-scale spatiotemporal learning through the Temporal Position Aggregator (TPA) and Spatial Sequence Aggregator (SSA). Together, these modules jointly capture latent dynamics and cross-domain dependencies in both sensor- and grid-based traffic scenarios.

\section{Conclusion}
We present STPFormer, a unified spatiotemporal forecasting framework that models complex dependencies across space and time. It combines a Temporal Position Aggregator (TPA), Spatial Sequence Aggregator (SSA), and Spatial-Temporal Graph Memory (STGM) to align temporal and spatial features more effectively, while an Attention Mixer fuses multi-scale information within the encoder. Unlike existing models with static encodings or separate blocks, STPFormer adapts dynamically to diverse traffic conditions. Experiments on five real-world datasets with varying traffic modes show that STPFormer consistently outperforms strong baselines for both sensor-based and grid-based forecasting, demonstrating its robustness and generalization.

\bibliography{main}  

\newpage
\section{Reproducibility Checklist}

This paper:

\begin{itemize}
\setlength{\itemsep}{3pt}
\setlength{\parsep}{0pt}
\setlength{\parskip}{0pt}
    \item Includes a conceptual outline and/or pseudocode description of AI methods introduced (yes/partial/no/NA)
    
    \textbf{Answer:}[\textcolor{blue}{Yes}]

    \textbf{Justification}: We have included a conceptual outline and pseudocode description of AI methods introduced.
    
    \item Clearly delineates statements that are opinions, hypothesis, and speculation from objective facts and results (yes/no)

    \textbf{Answer:}[\textcolor{blue}{Yes}]
    
    \textbf{Justification}: We have clearly delineates statements that are opinions, hypothesis, and speculation from objective facts and results.
    
    \item Provides well marked pedagogical references for less-familiare readers to gain background necessary to replicate the paper (yes/no)

    \textbf{Answer:}[\textcolor{blue}{Yes}]

    \textbf{Justification}: We have provides well marked pedagogical references for less-familiare readers to gain background necessary to replicate the paper.
\end{itemize}
Does this paper make theoretical contributions? (yes/no)

\noindent
\textbf{Answer:}[\textcolor{blue}{No}]

\noindent
If yes, please complete the list below.

\begin{itemize}
\setlength{\itemsep}{3pt}
\setlength{\parsep}{0pt}
\setlength{\parskip}{0pt}
    \item All assumptions and restrictions are stated clearly and formally. (yes/partial/no)
    
    \item All novel claims are stated formally (e.g., in theorem statements). (yes/partial/no)
    
    \item Proofs of all novel claims are included. (yes/partial/no)
    
    \item Proof sketches or intuitions are given for complex and/or novel results. (yes/partial/no)
    
    \item Appropriate citations to theoretical tools used are given. (yes/partial/no)
    
    \item All theoretical claims are demonstrated empirically to hold. (yes/partial/no/NA)
    
    \item All experimental code used to eliminate or disprove claims is included. (yes/no/NA)

\end{itemize}

\noindent
Does this paper rely on one or more datasets? (yes/no)

\noindent
\textbf{Answer:}\textcolor{blue}{Yes}

\noindent
\textbf{Justification}: We rely on more datasets.

\noindent
If yes, please complete the list below.

\begin{itemize}
\setlength{\itemsep}{3pt}
\setlength{\parsep}{0pt}
\setlength{\parskip}{0pt}
    \item A motivation is given for why the experiments are conducted on the selected datasets (yes/partial/no/NA)

    \textbf{Answer:}[\textcolor{blue}{Yes}]

    \textbf{Justification}: We have given the motivation why the experiments are conducted on the selected datasets.
    
    \item All novel datasets introduced in this paper are included in a data appendix. (yes/partial/no/NA)
    
    \textbf{Answer:}[\textcolor{blue}{NA}]

    \textbf{Justification}: This paper does not introduce any novel datasets; all experiments are conducted on publicly available datasets.
    
    \item All novel datasets introduced in this paper will be made publicly available upon publication of the paper with a license that allows free usage for research purposes. (yes/partial/no/NA)

    \textbf{Answer:}[\textcolor{blue}{NA}]

    \textbf{Justification}: This paper does not introduce any novel datasets; all experiments are conducted on publicly available datasets.
    
    \item All datasets drawn from the existing literature (potentially including authors’ own previously published work) are accompanied by appropriate citations. (yes/no/NA)

    \textbf{Answer:}[\textcolor{blue}{Yes}]

    \textbf{Justification}: We have cited appropriate citations of all datasets drawn from the existing literature (potentially including authors’ own previously published work).
    
    \item All datasets drawn from the existing literature (potentially including authors’ own previously published work) are publicly available. (yes/partial/no/NA)

    \textbf{Answer:}[\textcolor{blue}{Yes}]

    \textbf{Justification}: We have cited appropriate citations of all datasets drawn from the existing literature (potentially including authors’ own previously published work).
    
    \item All datasets that are not publicly available are described in detail, with explanation why publicly available alternatives are not scientifically satisficing. (yes/partial/no/NA)

    \textbf{Answer:}[\textcolor{blue}{No}]

    \textbf{Justification}:We do not have any undisclosed datasets.
\end{itemize}

\noindent
Does this paper include computational experiments? (yes/no)

\noindent
\textbf{Answer:}[\textcolor{blue}{Yes}]

\noindent
\textbf{Justification}: This paper includes computational experiments.
    
\noindent
If yes, please complete the list below.

\begin{itemize}
\setlength{\itemsep}{3pt}
\setlength{\parsep}{0pt}
\setlength{\parskip}{0pt}
    \item Any code required for pre-processing data is included in the appendix. (yes/partial/no).

    \textbf{Answer:}[\textcolor{blue}{Yes}]
    
    \item All source code required for conducting and analyzing the experiments is included in a code appendix. (yes/partial/no)

    \textbf{Answer:}[\textcolor{blue}{Yes}]

    \textbf{Justification}: We have included the code in the supplementary materials.
    
    \item All source code required for conducting and analyzing the experiments will be made publicly available upon publication of the paper with a license that allows free usage for research purposes. (yes/partial/no)

    \textbf{Answer:}[\textcolor{blue}{Yes}]

    \textbf{Justification}: We will make all source code required for conducting and analyzing the experiments publicly available upon publication of the paper with a license that allows free usage for research purposes.
    
    \item All source code implementing new methods have comments detailing the implementation, with references to the paper where each step comes from (yes/partial/no)

    \textbf{Answer:}[\textcolor{blue}{Yes}]

    \textbf{Justification}: We have comments detailing the implementation of all source code implementing new methods, with references to the paper where each step comes from.
    
    \item If an algorithm depends on randomness, then the method used for setting seeds is described in a way sufficient to allow replication of results. (yes/partial/no/NA)

    \textbf{Answer:}[\textcolor{blue}{Yes}]

    \textbf{Justification}: We have described the method used for setting seeds in a way sufficient to allow replication of results.
    
    \item This paper specifies the computing infrastructure used for running experiments, including GPU/CPU models; amount of memory; operating system; names and versions of relevant software libraries and frameworks. (yes/partial/no)

    \textbf{Answer:}[\textcolor{blue}{Yes}]

    \textbf{Justification}: We have specified the computing infrastructure used for running experiments, including GPU/CPU models; amount of memory; operating system; names and versions of relevant software libraries and frameworks.
    
    \item This paper formally describes evaluation metrics used and explains the motivation for choosing these metrics. (yes/partial/no)

    \textbf{Answer:}[\textcolor{blue}{Yes}]

    \textbf{Justification}: We have formally described the evaluation metrics used and explained the motivation for choosing these metrics. 
    
    \item This paper states the number of algorithm runs used to compute each reported result. (yes/no)

    Answer:[\textcolor{blue}{Yes}]

    \textbf{Justification}: We have stated the number of algorithm runs used to compute each reported result. 
    
    \item Analysis of experiments goes beyond single-dimensional summaries of performance (e.g., average; median) to include measures of variation, confidence, or other distributional information. (yes/no)

    \textbf{Answer:}[\textcolor{blue}{No}]

    \textbf{Justification}:While we did not explicitly report measures such as standard deviation or confidence intervals, our experimental evaluation spans multiple datasets, diverse traffic settings, and a variety of metrics (MAE, RMSE, MAPE). We also conducted comprehensive ablation studies and visual analyses (e.g., error heatmaps, module-wise outputs) to demonstrate robust and consistent improvements. These analyses help ensure that the observed performance gains are not limited to single conditions and reflect generalizable trends across different scenarios.
    \item The significance of any improvement or decrease in performance is judged using appropriate statistical tests (e.g., Wilcoxon signed-rank). (yes/partial/no)

    \textbf{Answer:}[\textcolor{blue}{No}]

    \textbf{Justification}: While we did not apply formal statistical significance tests such as the Wilcoxon signed-rank test, we conducted comprehensive ablation studies across multiple datasets and metrics. These consistently demonstrated the effectiveness of each proposed component and supported the robustness of the observed improvements.
    
    \item This paper lists all final (hyper-)parameters used for each model/algorithm in the paper’s experiments. (yes/partial/no/NA)

    \textbf{Answer:}[\textcolor{blue}{partial}]

    \textbf{Justification}: In the experimental section, we described our hyperparameters.
    
    \item This paper states the number and range of values tried per (hyper-) parameter during development of the paper, along with the criterion used for selecting the final parameter setting. (yes/partial/no/NA)

    \textbf{Answer:}[\textcolor{blue}{Partial}]
    
    \textbf{Justification}:  We report the final hyperparameter settings used in our experiments, including batch size, optimizer, learning rate, scheduler, early stopping, and seed.
\end{itemize}
\end{document}